\pdfoutput=1
\documentclass[11pt]{article}

\usepackage[]{emnlp2021}

\usepackage{times}
\usepackage{latexsym}

\usepackage[T1]{fontenc}
\usepackage[utf8]{inputenc}

\usepackage{microtype}

\usepackage{amsfonts,amsmath}
\usepackage{url}
\usepackage{paralist}
\usepackage{amsmath}
\usepackage{multirow}
\usepackage{multicol}
\usepackage{caption}
\usepackage{enumerate}
\usepackage{graphicx}
\usepackage{enumitem}
\usepackage{bm}
\usepackage[normalem]{ulem}
\usepackage{subfigure}
\usepackage{mathrsfs}
\usepackage{xcolor}

\title{Re-entry Prediction for Online Conversations \\via Self-Supervised Learning}

\author{Lingzhi Wang$^{1,2}$,
Xingshan Zeng$^3$, Huang Hu$^4$, Kam-Fai Wong$^{1,2}$, Daxin Jiang$^4$\\
	$^1$The Chinese University of Hong Kong, Hong Kong, China \\
	$^2$MoE Key Laboratory of High Confidence Software Technologies, China \\
	$^3$Huawei Noah’s Ark Lab, Hong Kong, China \\
	$^4$Microsoft Corporation, Beijing, China \\
	{ \tt $^{1,2}$\{lzwang,kfwong\}@se.cuhk.edu.hk}
	{ \tt $^3$zeng.xingshan@huawei.com} \\
	{ \tt $^4$\{huahu,djiang\}@microsoft.com} \\}
\date{}

\begin{document}
\maketitle
\begin{abstract}
In recent years, world business in online discussions and opinion sharing on social media is booming. 
Re-entry prediction task is thus proposed to help people keep track of the discussions which they wish to continue. 
Nevertheless, existing works only focus on exploiting chatting history and context information, and ignore the potential useful learning signals underlying conversation data, such as conversation thread patterns and repeated engagement of target users, which help better understand the behavior of target users in conversations. 
In this paper, we propose three interesting and well-founded auxiliary tasks, namely, Spread Pattern, Repeated Target user, and Turn Authorship, as the self-supervised signals for re-entry prediction. 
These auxiliary tasks are trained together with the main task in a multi-task manner.
Experimental results on two datasets newly collected from Twitter and Reddit show that our method outperforms the previous state-of-the-arts with fewer parameters and faster convergence. Extensive experiments and analysis show the effectiveness of our proposed models and also point out some key ideas in designing self-supervised tasks.\footnote{The code is available at \url{https://github.com/Lingzhi-WANG/ReEntryPrediction}}
\end{abstract}

\section{Introduction}
Online social media platforms are popular for individuals to discuss topics they are interested in and exchange viewpoints. However, a large number of online conversations are posted every day that hinder people from tracking the information they are interested in. As a result, there is a pressing demand for developing an automatic conversation management tool to keep track of the discussions one would like to keep engaging in.

Re-entry prediction \cite{zeng-etal-2019-joint,backstrom2013characterizing} is proposed to meet such demand. It aims to foresee whether a user (henceforth \textbf{target user}) will come back to a conversation they once participated in. Nevertheless, the state-of-the-art work \cite{zeng-etal-2019-joint} mostly focuses on rich information in users’ previous chatting history and ignores the thread pattern information \cite{backstrom2013characterizing,tan-etal-2019-context}. To this end, we study in re-entry prediction by exploiting the conversation thread pattern to signal whether a user would come back since the degree of repeated engagement of users can indicate their temporary interests in the ongoing conversation.

Self-supervised learning aims to train a model on labels that are automatically derived from the data itself. 
% , which attracts increasing research interests recently \cite{lan2019albert,erhan2010does,hinton2006fast}. 
Compared to previous generic self-supervised methods (e.g., Switch, Replace, and Mask), task-specific methods can achieve better performance \cite{jing2020self}, especially on medium-sized datasets, since task-oriented designs can better capture domain-specific features and thus achieve better performance for the target task. 
Therefore, we propose a prediction model (\textit{main model}) for re-entry prediction (\textit{main task}) with three auxiliary self-supervised tasks (\textit{Spread Pattern Prediction}, \textit{Repeated Target User Prediction} and \textit{Turn Authorship Prediction}) to assist learning of main model for re-entry prediction.

Spread Pattern Prediction is inspired by \textit{expansionary} and \textit{focused} thread in \citet{backstrom2013characterizing},
where thread pattern reflects the development of a conversation. 
We implement this task in a simplified but reasonable way to discriminate thread patterns based on the number of participated users. 
%  As we can see, the re-entry rate of focused thread is obviously higher than that of expansionary. Therefore, the thread patterns signal is
On the other hand, \citet{zeng-etal-2019-joint} shows that target users who contribute two or more posts in a conversation have a higher probability of coming back.
%We distinguish them in a more simplified way, to make our model aware of conversation's current state and therefore benefits further prediction.
% Conversations containing three or more users are labeled as expansionary and conversations with two users are labeled as focused. 
% The upper part of Figure 1 illustrate why our simplified from are reasonable. 
% On the other hand, \citet{zeng-etal-2019-joint} shows that target users who contribute two or more posts in a conversation have a larger probability of coming back.
% which is consistent with the results in Figure . 
% Therefore, we propose Repeated Target User Prediction task to benefit the training of main model by signaling target user's behaviour, i.e., whether target user has posted more than one messages in giving context.
Hence, we introduce a Repeated Target User Prediction task to facilitate the learning of the main model by capturing the target user's behavior, i.e., whether the target user has posted more than one message in a given context. 
% Finally, combining the aforementioned two tasks, we propose Turn Authorship Prediction. In this task, we step further from Repeated Target User task, to predict each turn's authorship.
Finally, we introduce the Turn Authorship Prediction task, in which we step further from the Repeated Target User Prediction task to predict if each turn's authorship is the target user. Thus, the model can track the participation of the target user and also know the thread pattern reflecting by the position of the target user who acts as a probe. 
\begin{figure}[t]
\centering
\includegraphics[width=0.45\textwidth]{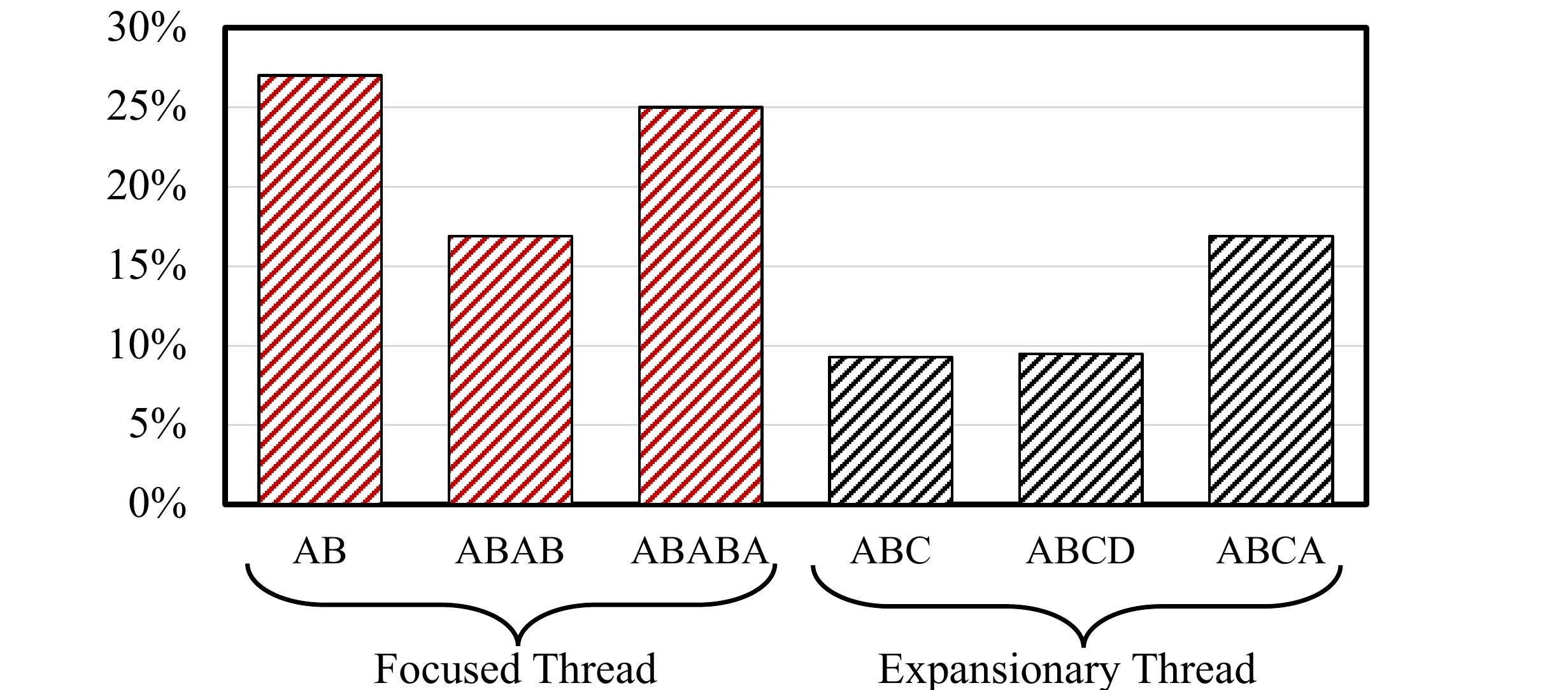}
% \vskip -0.5em
\caption{\label{fig:intro_case}  X-axis: thread pattern, e.g., ``AB'' represents thread where user A posts then B posts. Y-axis: re-entry rate, e.g., the re-entry rate for ``AB'' is 27\%, means that 27\% of the target users (user ``B'') in this kind of conversations will come back.
} 
% \vskip -1.5em
\end{figure}

To better illustrate our motivation, Figure \ref{fig:intro_case} shows the re-entry rate of six representative thread patterns on Reddit dataset. As we can see, the left three threads with user number ``$\leq 2$'' (focused) show a higher re-entry rate than the right three threads with user number ``$>2$'' (expansionary). We can also see that although ``ABCA'' is expansionary, it has repeated target user ``A'', which leads to a higher re-entry rate than the other two expansionary threads. Therefore, we can conclude that both Spread Pattern and Repeated Target User signals help predict re-entry behavior. Furthermore, since more challenging tasks get better performance \cite{mao2020survey}, we propose Turn Authorship Prediction, where we predict whether each turn's author is a target user or not.
%With the help of three devised self-supervised auxiliary tasks, our model can not only capture the conversation spread pattern but also track the participation trajectories of target user.
%In this way, we can not only capture target user's repeated engagement, but also know about how the conversation develops among the participants.

%Self-supervised tasks can combine these signals together and then help the learning of re-entry prediction task. Our goal is to solve Re-entry prediction (\textbf{Main Task}) while self-supervised tasks are auxiliary tasks where ground truth is obtained automatically. Many previous work has applied self-supervised learning, such as dialogue learning \cite{wu2019self}, text classification \cite{10.1007/978-3-030-55130-8_24} and story cloze test \cite{xu2020learning}. However, they just simply apply generic task, such as Mask, Switch and Replace etc. while our work design tasks by data itself.
Before the introduction of pretraining technique \cite{peters2018deep,devlin2018bert,radford2019language}, researchers focused on developing complex models \cite{lu2020conundrums}, such as key phrase generation with neural topic model \cite{wang2019topic} and structured models for coreference resolution \cite{martschat2015latent,bjorkelund2014learning}. Thus models are time-consuming in training and testing. For this reason, we propose our compact main model, which consists of three parts, turn encoder, conversation encoder and prediction layer. In addition, the chatting history information of the target user is also applied to our model by initializing the beginning hidden state of the target turn. The main model is jointly trained with the three self-supervised tasks in the manner of multi-task training and outperforms the BERT-based \cite{devlin2018bert} model which consists of large number of parameters and is time-consuming. 
%model  Our main model is simple, but training with the self-supervised tasks can achieve better performance than complex models.
%Therefore, we prefer small size but efficient models. 
%We propose a basic framework with several useful self-supervised task that can help training. The self-supervised task share most of the parameters with our main task and 

% In this work, our re-entry prediction model consists of turn encoder, conversation encoder and prediction layer.
% To jointly train the main model with proposed self-supervised tasks in a multi-task manner, we add a Multi-Layer Perceptron (MLP) on the top of model to predict the corresponding signal for each task.
% Besides, the chatting history information of target user is also applied to our model in the way of initialize the beginning hidden state of the target turn. 
% We conduct extensive experiments on two large-scale datasets collected from Reddit and Twitter. Results demonstrate the main model jointly training with auxiliary self-supervised tasks outperforms the previous state-of-the-arts.
In summary, our contributions are three-fold:
\begin{itemize}[leftmargin=*]
\setlength\itemsep{-0.2em}
\item Three self-supervised tasks are proposed to facilitate learning of the main model by capturing the thread pattern and participation trajectory of the target user.
\item Experimental results on two newly constructed datasets, Twitter and Reddit,  show that our methods outperform the previous state-of-the-arts with fewer parameters and faster convergence.
\item Extensive experiments and analyses provide more insights on how our models work and how to design effective self-supervised tasks for conversational prediction task. 
\end{itemize}
The remainder of this paper is organized as follows. The related work is surveyed in Section \ref{sec:related_work}. Section \ref{sec:model} and \ref{sec:self-sup} present the proposed approach, including model architecture and designed self-supervised tasks. Section \ref{sec:setup} and \ref{sec:results} then present the experimental setup and results respectively. Finally, conclusions are drawn in Section \ref{sec:conclusion}.
%Interesting analysis on why our self-supervised tasks work sheds light on how to design helpful self-supervised tasks.

%We propose learning the re-entry prediction model with multiple auxiliary self-supervised tasks to fully leverage the learning signals underlying conversation data.

%\item We devise the three auxiliary self-supervised tasks to enhance the capacity of re-entry prediction model in capturing the behaviours of target user in online conversations.

%\item Experimental results on two datasets collected from Twitter and Reddit show that our method outperforms the previous state-of-the-arts with fewer parameters and faster convergence.
% \item We propose three task-oriented self-supervised tasks and compare their performance with the existing self-supervised methods, such as Switch, Mask and Replace, etc. The experiments show that our model performs significantly better than the state-of-the-art and other  self-supervised baselines.
% \item We study convergence rate and robustness of the model (sensitivity to the initialization of model parameters and the size of datasets).

\section{Related Work}\label{sec:related_work}
%Our work is in line with re-entry prediction and self-supervised learning.
\paragraph{Re-entry Prediction.} Re-entry prediction \cite{zeng-etal-2019-joint,backstrom2013characterizing,budak2013participation} aims to forecast whether the users will return to a discussion they once entered and \citet{zeng-etal-2019-joint} achieves state-of-the-art performance by exploiting user’s history context \cite{flek2020returning}. Re-entry prediction focuses on conversation-level response prediction \cite{zeng2018microblog,chen2011speak}. Most of them adopt a complex framework \cite{zeng2018microblog} and massive parameters (see Figure \ref{sfig:exp:size_time}) while our model is simple and effectively combines the current conversation and chatting history. 
\paragraph{Self-supervised Learning.} Self-supervised learning aims to train a network on an auxiliary task where the ground-truth label is automatically derived from the data itself \cite{wu2019self,lan2019albert,erhan2010does,hinton2006fast}. It has been applied to many tasks, such as text classification \cite{yu2016learning}, neural machine translation \cite{ruiter-etal-2019-self}, multi-turn response selection \cite{xu2020learning}, summarization \cite{chen2019sequential} and dialogue learning \cite{wu2019self}. These auxiliary tasks can be categorized into word-level tasks and sentence-level tasks. 

In word-level tasks, nearby word prediction  \cite{mikolov2013distributed} and next word prediction  \cite{bengio2003neural,wang2015unsupervised} are widely explored in language modeling. Masked language model \cite{devlin2018bert} is also in the line of word-level tasks. 

In sentence-level tasks, \citet{wang2019self} exploits Mask, Replace and Switch for extractive
summarization. \citet{wu2019self} propose Inconsistent Order Detection for dialogue learning. \citet{10.1007/978-3-030-55130-8_24} exploit Drop, Replace, and TOV (Temporal Order Verification) for story cloze test. \citet{xu2020learning} also design several self-supervised tasks to improve the performance of response selection.

Most of the previous self-supervised tasks (both in word-level and sentence-level) focus on the general domain while our work is based on task-orientated supervised methods and achieves better performance. 
%\citet{wu2019self} proposes the inconsistent order detection task to explicitly capture the order information and to help the training of the main task. \citet{10.1007/978-3-030-55130-8_24} has three auxiliary tasks: Drop, Replace and TOV (Temporal Order Verification). \citet{xu2020learning} designs several self-supervised tasks (next session prediction, utterance restoration, incoherence detection and consistency discrimination.) to improve the multi-turn response selection. 
%The differences between them is adopting these technologies into different setting of task.
%Some of them evaluate different auxiliary tasks' performance but most of them concentrate on F1-like metrics and neglect model's stability and convergence speed, which are essential. Our work does not only evaluate six self-supervised methods (the most comprehensive so far) respectively and combinedly but also concentrate on the un-functional performance.
%Two related confusing concepts, self-attention and self-learning is out of the scope of our paper. 
\section{Re-entry Prediction Framework}\label{sec:model}
\begin{figure*}[t]
\centering
\includegraphics[width=1.\textwidth]{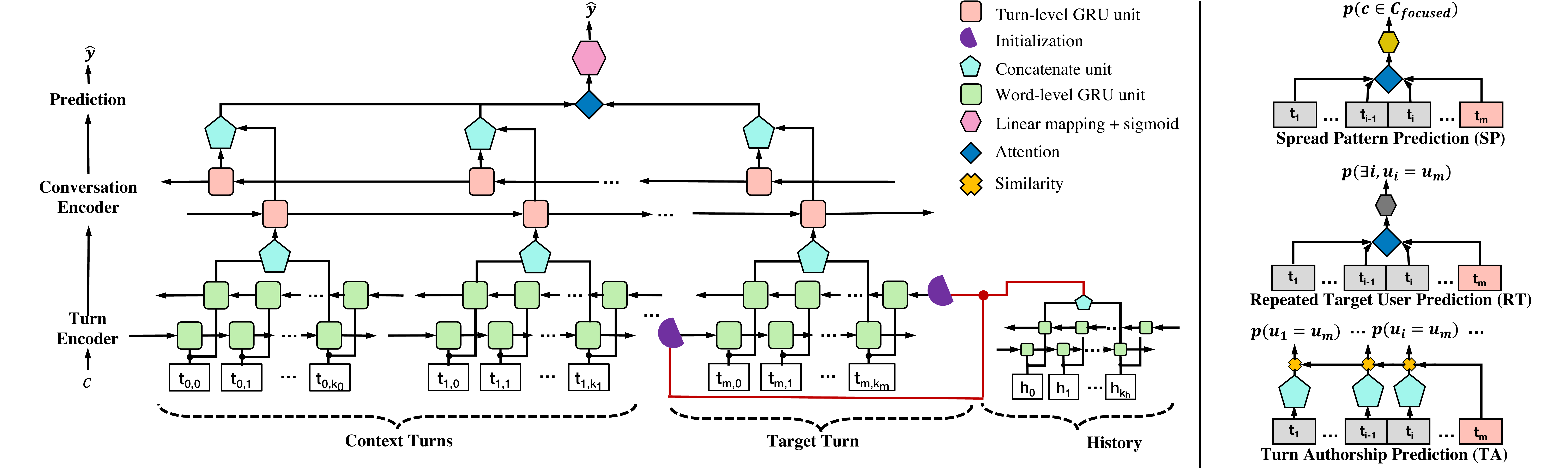}
\caption{\label{fig:sketch} Our main model (left part) and three self-supervised tasks (right part) for re-entry prediction. }
\vskip 1em
\end{figure*}
This section describes our re-entry prediction framework. The left part of Figure \ref{fig:sketch} shows our overall structure. In the following, we first introduce the input and output in Section \ref{ssec:model:input_output}. Then in Section \ref{ssec:model:predict}, we describe our prediction framework. 
Finally, the learning objective of the entire model will be given in Section \ref{ssec:model:learning-obj}.

\subsection{Input and Output}\label{ssec:model:input_output}
The input of our model contains two parts: the observed conversation $c$ and chatting history $c^h$ of target user $u$. 
The conversation $c$ is formalized as a sequence of turns (e.g., posts or tweets) $<t_1, t_2, ..., t_m>$ where $m$ represents the length of conversation (number of turns) and $t_m$ is posted by user $u$. 
$t_i$ is the $i$-th turn of the conversation and contains words $<w_{i,1}, w_{i,2}, ..., w_{i,k_i}>$, where $w_{ij}$ is the $j$-th word in $i$-th turn and $k_i$ is the word length of $i$-th turn. The chatting history $c^h$ is constructed by concatenating the turns (in training corpus) that are authored by the user $u$ into a sequence following their posting time. 
%Each history turn has similar structure as the turns in the observed conversation $c$.

For output, we yield a Bernoulli distribution $p(c, c^h)$ to indicate the estimated likelihood of whether $u$ will re-engage in
the conversation $c$, giving the chatting history $c^h$ of $u$.

\subsection{Re-entry Prediction Model}\label{ssec:model:predict}
Our model consists of three modules: turn encoder, conversation encoder, and prediction layer.
\paragraph{Turn Encoder.}
We first feed each word $w_{i,j}$ in turn $t_i$ into an embedding layer and get the word representation $\bm{e}_{i,j}$. 
% Similarly, we get word representation $e^h$ for chatting history $c^h$.
The turn representation is then modeled via a turn encoder, where a word-level bidirectional gated recurrent unit (Bi-GRU) \cite{DBLP:conf/emnlp/ChoMGBBSB14} is adopted. The hidden states of Bi-GRU are defined as:
% \vskip -0.8em
\begin{equation}\small \label{equ:turn-gru}
\overrightarrow{\bm{h}_{j}} = f_{GRU}(\bm{e}_{j}, \bm{h}_{j-1}), \,\overleftarrow{\bm{h}_{j}} = f_{GRU}(\bm{e}_{j}, \bm{h}_{j+1})
\end{equation}
% \vskip -0.2em
\noindent The output of our turn encoder is the concatenation of the last hidden states of both directions of Bi-GRU: $\bm{h}= [\overrightarrow{\bm{h}_{j}};\overleftarrow{\bm{h}_{j}}]$.

To incorporate the information of the user's chatting history, we also use the same procedure described above to encode each history turn $t_i^h$ in $c^h$. We then apply another Bi-GRU layer to capture the temporal features among these history turns and derive the final representation of chatting history $\bm{h}^h$ for target user $u$.
Finally, we use $\bm{h}^h$ to initialize the hidden states of the last conversation turn (posted by target user $u$) $t_m$'s turn encoder, and the initialization mechanism is proven to be helpful in \citet{wang-etal-2020-continuity}.
% by Bi-GRU and this Bi-GRU is initialized by the representation of chatting history $\bm{h}^h$. 
The following equation describes the initialization: 
% \vskip -0.8em
\begin{equation}\small \label{equ:init}
\overrightarrow{\bm{h}_{m,0}} = \overleftarrow{\bm{h}_{m,k_m}} = \eta (\bm{W}_0 \bm{h}^h+b_0)
\end{equation}
% \vskip -0.2em
\noindent where $\overrightarrow{\bm{h}_{m,0}}$ and $\overleftarrow{\bm{h}_{m,k_m}}$ are the initial states of both directions, and $\eta$ is a Tanh activated function. $\bm{W_0}$ and $b_0$ are learnable parameters.

% Concretely, the initial states for both direction are $\overrightarrow{\bm{h}_{m,0}} = \overleftarrow{\bm{h}_{m,k_m}} = \bm{W}_0 h^h+b_0 $, where $\bm{W_0}$ and $b_0$ are learnable parameters. Then we concatenate the two directions of Bi-GRU to get the representation $h_m$ of turn $t_m$. We can also get representations of the turns before the last turn $<h_0, h_1, ..., h_{m-1}>$ following the scheme of modeling  chatting history.

With the initialization process, we produce more informative representation of the final turn $\bm{h}_m$, since it can encode information from both current conversation $c$ and target user $u$'s chatting history.
\paragraph{Conversation Encoder.}
To learn the conversational structure representations for $c$, we apply a third Bi-GRU, to capture the interactions between adjacent context turns: 
% \vskip -0.8em
\begin{equation}\small \label{equ:conv-gru}
\overrightarrow{\bm{r}_{j}} = f_{GRU}(\bm{h}_{j}, \bm{r}_{j-1}), \,\overleftarrow{\bm{r}_{j}} = f_{GRU}(\bm{h}_{j}, \bm{r}_{j+1})
\end{equation}
% \vskip -0.2em
\noindent We concatenate the outputs of both directions and get the turn representations of $c$: $\bm{r}_j = [\overrightarrow{\bm{r}_{j}};\overleftarrow{\bm{r}_{j}}]$.

Since different turns might play different roles in predicting target user $u$'s re-entry behavior (e.g. the turns that $u$ directly replied before should be more important than other turns), we apply an attention mechanism here to force our model to pay more attention to important turns. Concretely, the final representation of conversation $c$ is defined as:
% \vskip -0.8em
\begin{equation}\small \label{equ:att}
\bm{r} = \sum_{j=0}^m a_j \bm{h}_j, \, a_j = softmax (\bm{W}_{att} \bm{h}_j + b_{att})
\end{equation}
% \vskip -0.2em
\noindent where $\bm{W}_{att}$ and $b_{att}$ are learnable parameters.
\paragraph{Prediction Layer.} We predict the final output $\hat{y} \in [0,1]$, which signals how likely $u$ will re-engage in $c$, with the following prediction layer:
% \vskip -0.8em
\begin{equation}\small \label{equ:predict}
\hat{y} = \sigma (\bm{v}^T [\bm{r}, \bm{h}_m] + b)
\end{equation}
% \vskip -0.2em
\noindent where $\bm{v}$ and $b$ are learnable parameters and $\sigma()$ is the sigmoid function. Here we concatenate the conversation representation $\bm{r}$ and hidden state of final turn $\bm{h}_m$ as input, to emphasize the role of final turn posted by target user $u$.

\subsection{Learning Objective}\label{ssec:model:learning-obj}
Following~\citet{zeng-etal-2019-joint}, we use \emph{binary cross-entropy loss} as our learning objective.
Also, to deal with the imbalance of positive and negative instances in training corpus, we weigh differently for their losses. The equation for our main task is defined as follows:
% \vskip -1.5em
\begin{equation} \label{equ:loss}
\small
\mathcal{L}_{main} = - \sum_{i\in \mathcal{T}} \big [ \lambda \cdot y_i \log (\hat{y_i}) + \mu (1 - y_i) \log (1 - \hat{y_i}) \big ]
\end{equation}
% \vskip -0.5em
\noindent where $\mathcal{T}$ is the training corpus, 
$y_i$ denotes the ground-truth label for $i$-$th$ instance in the training corpus (label is $1$ if target user re-engages later, otherwise $0$), and $\lambda$ and $\mu$ are hyper-parameters to trade off the weights between positive and negative instances.
Generally, the values for $\lambda$ and $\mu$ can be tuned based on the ratios of positive and negative examples in the training corpus.

\section{Self-Supervised Tasks}\label{sec:self-sup}
This section describes the proposed self-supervised tasks that guide the re-entry prediction model to better capture user behaviors in online conversations.
% These auxiliary tasks are designed to capture the relevance, coherence, and consistency in conversation turns, and reveal user $u$'s habitual behavior in online conversations. 
The right part of Figure \ref{fig:sketch} illustrates our three self-supervised tasks. 
% We introduce them in turn as follows.

\subsection{Spread Pattern Prediction}
\citet{backstrom2013characterizing} shows that \textit{expansionary} (engagement among many users) and \textit{focused} (repeated engagement among few users) are two kinds of spread patterns in online multi-party conversations. Distinguishing spread patterns of conversations is helpful in predicting the future trajectory of the conversation. Therefore, we propose the \textit{Spread Pattern Prediction} task (\textbf{SP} task in Figure~\ref{fig:sketch}) which is a simplified form of the work of \citet{backstrom2013characterizing}.

We divide conversations into two types --  \textit{focused conversation} ($C_{focused}$) and \textit{expansionary conversation} ($C_{exp}$). Focused conversations are composed of repeated discussions between only two active users while expansionary conversations contain more than two users. We then make binary prediction between focused (label $y^{sp}=1$) and expansionary ($y^{sp}=0$) conversation with another prediction layer (the reason for assigning label $1$ to focused conversation can be found in Section \ref{ssec:why}):
% \vskip -1.5em
\begin{equation}\small \label{equ:predict-SP}
\hat{y}^{sp} = p(c \in C_{focused}) = \sigma (\bm{v}_{sp}^T [\bm{r}, \bm{h}_m] + b_{sp})
\end{equation}
% \vskip -0.5em
\noindent where $\bm{r}$ and $\bm{h}_m$ are the same as Eq.~\ref{equ:predict}, $\bm{v}_{sp}$ and $b_{sp}$ are learnable parameters.

We still apply weighted binary cross entropy introduced in Eq.~\ref{equ:loss} as our learning objective. To simplify the hyper parameter tuning, we force the trade off weight to be the ratio between positive and negative instances. Below describes the equation: 
% \vskip -1em
\begin{equation} \label{equ:loss-SP}
\small
\mathcal{L}_{SP} = - \sum_{i\in \mathcal{T}} \big [ \lambda_{sp} \cdot y_i^{sp} \log (\hat{y}_i^{sp}) + (1 - y_i^{sp}) \log (1 - \hat{y}_i^{sp}) \big ]
\end{equation}
% \vskip -0.5em
\noindent where $\lambda_{sp}$ equals to the number of positive instances divided by that of negative ones in training corpus. $\hat{y}_i^{sp}$ is the output of the $i$-th instance.

%In \citet{backstrom2013characterizing}, authors show that there might be two kinds of long comment threads -- \textit{expansionary} and \textit{focused} threads. The former ones are referred to as those attract many different commenters to contribute, while the latter ones come from repeated engagement among relatively few users. Understanding their spread patterns is useful for characterizing and predicting conversation threads. Based on such idea, we design our first self-supervised task as \textit{Spread Pattern} (SP) Prediction. To simplify the prediction, we divide our target conversations into two types. Those belonging to the first type are formed with repeated discussions between only two active users, where we mark them as ``focused'' conversation. The other type of conversations contain more than two users, and so they are more ``expansionary'' conversations. 

%We assume that users in a ``focused'' conversation are more likely to return, since they have already repeatedly contributed to that conversation and therefore might be more interested in its future trend. What's more, by distinguishing these two kinds of conversations, we can learn more expressive representations for the conversation context turns.
%To formalize the task, we label the ``focused'' conversations with $1$, and the others with $0$. Then our model can do the binary prediction under such supervision (A task in Figure~\ref{fig:sketch}).

\subsection{Repeated Target Prediction}
\citet{zeng-etal-2019-joint} shows that their model achieves better performance in second or third re-entry prediction (i.e. the target user has already contributed two or three turns) than first re-entry prediction. It might be attributed to the fact that users who participated in the conversation twice or more are more likely to return to this conversation (see statistic in Table \ref{statistics-table1}). 
%With such observation, we believe that repeated target user can be a signal to help prediction in main task.
Therefore, we design \textit{Repeated Target Prediction} task (refer to \textbf{RT} task in Figure~\ref{fig:sketch}). We label those conversations containing repeated target users with $1$ ($y^{rt}=1$) and other conversations with $0$ ($y^{rt}=0$) and carry out binary prediction:
% \vskip -1.5em
\begin{equation}\small \label{equ:predict-RT}
\hat{y}^{rt} = p(\exists i, u_i = u_m) = \sigma (\bm{v}_{rt}^T [\bm{r}, \bm{h}_m] + b_{rt})
\end{equation}
% \vskip -1em
\noindent where $\bm{r}$ and $\bm{h}_m$ are the same as Eq.~\ref{equ:predict}, $\bm{v}_{rt}$ and $b_{rt}$ are learnable parameters.

The learning objective for this task can be summarized as below, following similar idea in SP:
% \vskip -1.5em
\begin{equation} \label{equ:loss-RT}
\small
\mathcal{L}_{RT} = - \sum_{i\in \mathcal{T}} \big [ \lambda_{rt} \cdot y_i^{rt} \log (\hat{y}_i^{rt}) + (1 - y_i^{rt}) \log (1 - \hat{y}_i^{rt}) \big ]
\end{equation}
% \vskip -4em

\subsection{Turn Authorship Prediction}
By combining the intention of the SP and RT tasks, we further design \textit{Turn  Authorship Prediction} (henceforth \textbf{TA}) task. The TA task aims to predict whether the turn's author is the target user and we label "yes" with $1$ and "no" with $0$. This task benefits the main task by signaling both the conversation spread pattern and repeated user pattern. Specifically, this is a turn-level authorship prediction and can help learn meaningful turn representations, which are essential for conversation modeling. 

Formally, we output each turn's score as the similarity between hidden states of current turn ($\bm{h}_i$, $i = 1, 2, ..., m-1$) and target turn ($\bm{h}_m$), followed by a sigmoid activated function:
% \vskip -1em
\begin{equation}\small \label{equ:predict-TA}
\hat{y}^{ta}_j = p(u_j = u_m) = \sigma (\bm{h}_j\cdot\bm{h}_m)
\end{equation}
% \vskip -0.5em
\noindent which reflects the probability of turn $j$ being authored by the target user. Mean Square Error (MSE) loss is applied for TA task:
% \vskip -1em
\begin{equation} \label{equ:loss-TA}
\small
\mathcal{L}_{TA} = \sum_{i\in \mathcal{T}} \sum_{j=1}^{m_i-1} (y_{ij}^{ta} - \hat{y}_{ij}^{ta})^2
\end{equation}
% \vskip -0.5em
\noindent where $m_i$ is the turn number of conversation $i$.

% We combine our thoughts in first and second tasks into one target, as our third self-supervised task, where we not only predict the developing patterns of current conversation but also predict whether target user has repeatedly appeared in the conversation. To that end, we predict whether the author of each turn in conversations is the target user, named \textit{Turn  Authorship} (TA) Prediction.
% Specifically, we label the turns authored by target user with $1$, and other turns with $0$.
% Then our model will make the turn-level prediction (C task in Figure~\ref{fig:sketch}).
% With this task, our model learns more meaningful turn-level representations, which benefit later prediction. 

\subsection{Training Procedure}
All three auxiliary tasks are trained on parameters shared with the main task except for the final prediction layer (Section~\ref{ssec:model:predict}) in a multi-task learning manner. 
The final total loss is:
% \vskip -1.5em
\begin{equation} \label{equ:loss-final}
\small
\mathcal{L}_{final} = \mathcal{L}_{main} + \alpha_{sp}\mathcal{L}_{SP} + \alpha_{rt}\mathcal{L}_{RT} + \alpha_{ta}\mathcal{L}_{TA}
\end{equation}
% \vskip -1em
\noindent where $\alpha_{sp}$, $\alpha_{rt}$, $\alpha_{ta}$ are hyper-parameters.
%between main task and the other tasks. 
% We try both self-supervised pre-training and multi-task learning for all three tasks. Then we compare their effects in enhancing re-entry prediction. Detailed results will be showed in Section ?.

\section{Experimental Setup}\label{sec:set}
\label{sec:setup}
\paragraph{Datasets.} For experiments, we construct two new datasets from \textbf{Twitter} and \textbf{Reddit}. The raw Twitter and Reddit data is released by 
% \citet{wang-etal-2020-continuity} and Reddit obtained from
\citet{zeng2018microblog,zeng-etal-2019-joint} and both in English. For both Twitter and Reddit, we form the conversations with postings and replies (all the comments and replies also viewed as a single turn) following the practice in \citet{DBLP:conf/emnlp/LiGWPW15} and \citet{zeng2018microblog}. 

In our main experiment, different from \citet{zeng-etal-2019-joint}, we do not focus on predicting first re-entries (i.e. only giving the context until the target user's first participation), we generalize the setting into re-entry prediction regardless of the number of user's past participation. In this way, our model can learn more general and applicable features for re-entry prediction in diverse scenarios.

\begin{table}[t]\footnotesize
\newcommand{\tabincell}[2]{\begin{tabular}{@{}#1@{}}#2\end{tabular}}
\begin{center}
\begin{tabular}{|l|rr|}
\hline 
&  \textbf{Twitter} & \textbf{Reddit} \\
\hline
{\# of convs} & 45,111 & 16,340  \\
{\# of turns} & 229,435 & 58,189  \\
{Avg \# of turns per conv}& 5.09 & 3.56  \\
{Avg len of turn per conv}&20.3 &42.9 \\
%{$|$Voc$|$ of convs}& 43822 & 70414  \\
% {$|$Voc$|$ of convs}& 88,867 & 46,494  \\
{\% with repeated target}& 63.2 & 39.7  \\
{\% of positive instances}& 48.9 & 21.3 \\
\hline
\end{tabular}
\end{center}
% \vskip -1em
 \caption{
 \label{statistics-table1}Statistics of Twitter and Reddit datasets. 
%  The upper rows are for quotes and the lower rows are for conversations.
 ``Avg \# of turns'' means the average turn number. ``len'' refers to the number of tokens. ``\% with repeated target'' represents the ratio of conversations that target users have appeared at least twice in context. ``Positive instances'' are the conversations which target users re-engage later.
% Avg len: average number of tokens. Avg \# of turns per conv: average number of turns in a conversation. 
%$|$Voc$|$: the vocabulary size. 
}
% \vskip -1em
\end{table}
The statistics of the two datasets are shown in Table \ref{statistics-table1}. As can be seen, Twitter dataset is much larger than Reddit dataset, with longer conversations  (derived from the average number of turns)
and shorter turns (observed from the average length of turns).
Besides, it contains more conversations with repeated target users, which means that we are more likely to predict the second or third re-entries. At last, Reddit dataset is severely imbalanced in the ratio of positive and negative samples. This indicates that users in Reddit usually do not come back to the conversation they once participated in. 
%The two datasets show very different characteristics, so that we can test our method in different scenarios.

\begin{figure}[t]
\centering
\includegraphics[width=0.45\textwidth]{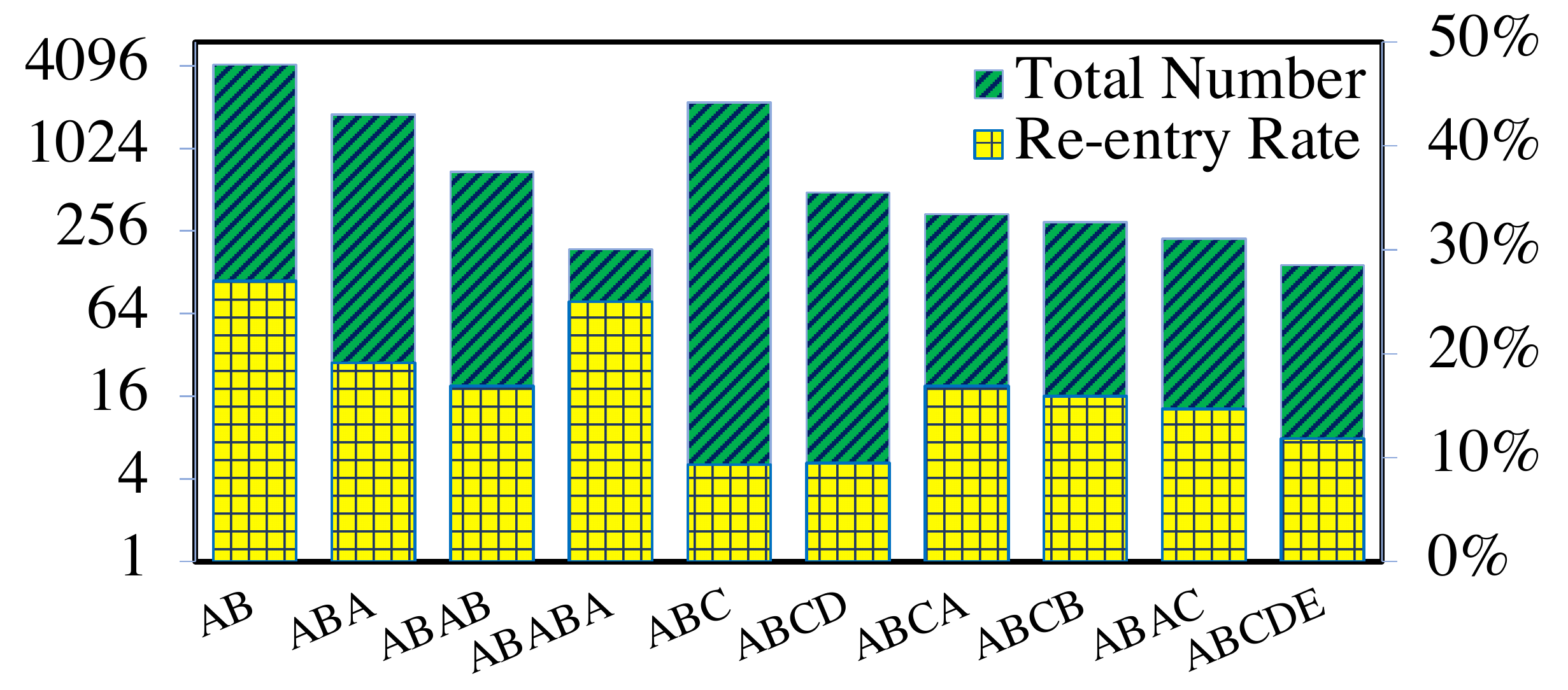}
% \vskip -1em
\caption{\label{fig:statistis:pattern_rate}  X-axis: thread patterns (the same meaning as in Fig. \ref{fig:intro_case}). 
%For example, ``ABA" refers to the conversations posted by user A and B, and they post in the order of ``ABA".
Left Y-axis: the number of user patterns; Right Y-axis: re-entry rate for each user pattern. 
%For example, the re-entry rate for conversations with pattern ``ABA" is 19\%, means that 19\% of the target users in these conversations will come back.
} 
% \vskip -1em
\end{figure}

We also present the distribution of thread patterns with their re-entry rate for Reddit in Figure \ref{fig:statistis:pattern_rate}. 
%User pattern means the order that users post. 
It can be seen that ``AB'', ``ABA'' and ``ABC'' are the most frequent patterns. 
%, which means that first or second re-entries are most important cases for prediction. 
And re-entry rate for focused conversations (i.e. only two users participate, such as ``AB'' and ``ABAB'') is generally higher than expansionary conversations, since prior contributions in one conversation may result in continued participation. 
Such a phenomenon verifies our motivation to design self-supervised tasks. 

\paragraph{Preprocessing.}
We applied the Glove tweet preprocessing toolkit \cite{pennington2014glove} to the Twitter dataset. As for the Reddit dataset, we first tokenized the words with the open-source natural language toolkit (NLTK) \cite{10.3115/1118108.1118117}. We then removed all the non-alphabetic tokens and replaced links with the generic tag ``URL''. For both datasets, a vocabulary was maintained with all the remaining tokens, including emoticons and punctuation marks.

\paragraph{Parameter Setting.}
For the parameters in the main model, we first initialize the embedding layer with 200-dimensional Glove embedding \cite{pennington2014glove}, whose Twitter version is used for the Twitter dataset and Common Crawl version is applied to Reddit\footnote{\url{https://nlp.stanford.edu/projects/glove/}}. For our BiGRU layers, we set the size of hidden states for each direction to $200$. We employ Adam optimizer~\cite{kingma:adam} with initial learning rate $1e$-$4$ and early stop adoption~\cite{caruana2001overfitting} in training. The batch size is set to 32.  
Dropout strategy~\cite{Srivastava:2014:DSW:2627435.2670313} and $L_2$ regularization are used to alleviate overfitting. And the tradeoff parameters $\alpha_{sp}$, $\alpha_{rt}$, $\alpha_{ta}$ are all set to $0.2$. All the hyper-parameters above are tuned on the validation set by grid search.

\paragraph{Evaluation Metrics.}
% To evaluate the performance, evaluation metrics for binary classification, 
% including area under the ROC Curve (AUC), accuracy (ACC), precision (Pre), and F1-scores (F1) are adopted.
We use area under the Curve of ROC (AUC), accuracy (ACC), precision (Pre), and F1-scores (F1) to evaluate baselines and our method.
Note that, to save spaces, we do not include Recall scores since it can be calculated with Pre and F1.

\paragraph{Baselines and Comparisons.}
We first compare four baselines.
The first method is a weak baseline \underline{\textsc{Random}} that randomly predicts "yes-or-no" labels. 
The second model, referred to as \underline{\textsc{CCCT}}, is from an earlier work~\cite{backstrom2013characterizing} that trains a bagged decision tree with manually-crafted features including arrival patterns, time effects, and most related terms, etc.
The third compared model, \underline{\textsc{BiLSTM+BiA}} \cite{zeng-etal-2019-joint}, yields state-of-the-art results with a BiLSTM modeling turn information and a bi-attention mechanism extracting the interaction effects between context and history. 
We also compare \underline{\textsc{BERT+BiLSTM}}, where the turn representations are extracted with BERT~\cite{devlin2018bert}, and a BiLSTM is applied for modeling the conversation structure. For our proposed method, we further compare different self-supervised tasks (SP, RT, TA). 
%for both pre-training and multi-task learning scenarios. The three tasks (SP, RT, TA) are evaluated independently with both pre-training and multi-task learning, followed by their joint effects for final prediction.

% 1)Successful  
% We also study the effectiveness of history by varying the use of history information.
\section{Experimental Results}\label{sec:results}
In this section, we first introduce the main comparison results in Section \ref{ssec:exp:main_results}. Then the effects of our methods and how our methods make effects are given in Section \ref{ssec:comparison_in_detail} and Section \ref{ssec:why} respectively. 
Finally, Section \ref{ssec:more_discuss} yields further discussion on user history and error analysis.
%Then we show the effects of our methods in detail by comparing with other methods in Section \ref{ssec:comparison_in_detail}. We also show how our methods make effects in Section \ref{ssec:why} and more discussion are given in Section \ref{ssec:more_discuss}.

\subsection{Main Comparison Results} \label{ssec:exp:main_results}
\begin{table*}[t]\setlength{\tabcolsep}{1.2mm}\small
\newcommand{\tabincell}[2]{\begin{tabular}{@{}#1@{}}#2\end{tabular}}
\begin{center}
\scalebox{0.9}{
\begin{tabular}{|l|rrrr|rrrr|}
\hline 
\multirow{2}{*}{Models} 
&\multicolumn{4}{c|}{ \tabincell{c}{\textbf{Twitter}} }
&\multicolumn{4}{|c|}{ \tabincell{c}{\textbf{Reddit}} }

\\
\cline{2-9}
& AUC  & Acc & Pre & F1 & AUC &  Acc & Pre & F1 \\
\hline
\hline
\underline{\textbf{Baselines}} &&& & & & & &\\
\textsc{Random}& 50.3\scriptsize{$\pm0.56$} & 50.1\scriptsize{$\pm0.54$} &  51.2\scriptsize{$\pm0.55$}&  50.5\scriptsize{$\pm0.55$}
& 49.3\scriptsize{$\pm1.38$} & 49.5\scriptsize{$\pm1.44$} & 22.0\scriptsize{$\pm1.24$} &  30.6\scriptsize{$\pm1.73$}\\
\textsc{CCCT}\cite{backstrom2013characterizing} & 62.4 & 60.3 & 57.9  &64.9
& 60.1 & 57.2 & 29.5 &36.1\\
\textsc{BiLSTM+BiA}\cite{zeng-etal-2019-joint}& 57.3\scriptsize{$\pm1.18$} & 51.8\scriptsize{$\pm0.45$} &52.3\scriptsize{$\pm0.21$} &  67.9\scriptsize{$\pm0.19$}
& 60.9\scriptsize{$\pm2.75$} & 55.8\scriptsize{$\pm3.81$} & 28.1\scriptsize{$\pm1.70$}  &38.3\scriptsize{$\pm2.08$}\\
\textsc{BERT+BiLSTM}\cite{devlin2018bert}& 67.8\scriptsize{$\pm0.26$} & 60.8\scriptsize{$\pm0.33$} & 57.2\scriptsize{$\pm0.22$} & 69.7\scriptsize{$\pm0.19$}
& 62.5\scriptsize{$\pm0.11$} & 55.3\scriptsize{$\pm2.00$} & 28.2\scriptsize{$\pm1.06$} & 39.5\scriptsize{$\pm0.33$}\\
\hline
\hline
\underline{\textbf{W/O Self-supervised Task}} & & &&& & & & \\
\textsc{BiGRU}& 65.4\scriptsize{$\pm0.69$} & 58.1\scriptsize{$\pm1.59$} & 54.9\scriptsize{$\pm1.55$} &  67.8\scriptsize{$\pm0.35$} 
& 58.6\scriptsize{$\pm3.00$} & 48.1\scriptsize{$\pm3.85$} & 26.1\scriptsize{$\pm1.87$} &  38.1\scriptsize{$\pm1.24$}\\
\textsc{BiGRU+History}& 65.1\scriptsize{$\pm0.64$} &58.2\scriptsize{$\pm1.21$} & 55.2\scriptsize{$\pm1.13$} &  68.6\scriptsize{$\pm0.53$}
& 61.8\scriptsize{$\pm3.15$} & 52.4\scriptsize{$\pm2.42$} &27.4\scriptsize{$\pm1.63$}  &39.1\scriptsize{$\pm1.46$}\\
\textsc{BiGRU+Att}& 66.5\scriptsize{$\pm0.79$} &59.3\scriptsize{$\pm1.11$} & 56.2\scriptsize{$\pm1.14$} &  68.7\scriptsize{$\pm0.40$}
&59.3\scriptsize{$\pm3.95$} & 51.7\scriptsize{$\pm3.39$} & 27.1\scriptsize{$\pm2.50$}  &37.5\scriptsize{$\pm1.35$}\\
\textsc{BiGRU+His+Att(Full Main)}& 67.3\scriptsize{$\pm0.62$} &59.9\scriptsize{$\pm1.39$} & 56.7\scriptsize{$\pm1.15$} &  69.4\scriptsize{$\pm0.83$}

& 61.6\scriptsize{$\pm3.93$} & 53.4\scriptsize{$\pm4.11$} & 29.1\scriptsize{$\pm2.93$} &39.4\scriptsize{$\pm1.51$}\\
\hline
\hline
\underline{\textbf{With Self-supervised Task(s)}} & & &&& & & &\\
\textsc{Full Main+SP}& 67.1\scriptsize{$\pm0.47$} & 59.9\scriptsize{$\pm1.10$} & 57.1\scriptsize{$\pm1.04$} & 69.9\scriptsize{$\pm0.30$} 
& 62.8\scriptsize{$\pm0.82$}
 & 58.1\scriptsize{$\pm2.18$} &29.6\scriptsize{$\pm1.42$} &40.0\scriptsize{$\pm0.25$}\\
 
\textsc{Full Main+RT}& 67.4\scriptsize{$\pm0.41$} & 60.0\scriptsize{$\pm1.06$} & 57.1\scriptsize{$\pm1.02$} & 69.3\scriptsize{$\pm0.16$} 
 & 63.2\scriptsize{$\pm1.40$} & \textbf{59.6}\scriptsize{$\pm1.86$} & \textbf{30.1}\scriptsize{$\pm1.14$} &39.9\scriptsize{$\pm0.98$}\\
 
 \textsc{Full Main+TA}& \textbf{68.6}\scriptsize{$\pm0.86$} &\textbf{61.0}\scriptsize{$\pm0.90$} & \textbf{58.4}\scriptsize{$\pm0.91$} & \textbf{70.5}\scriptsize{$\pm0.19$} 
 & \textbf{64.6}\scriptsize{$\pm0.91$} & 57.7\scriptsize{$\pm2.12$} &29.1\scriptsize{$\pm1.81$} &\textbf{40.6}\scriptsize{$\pm0.27$}\\
 
% \textsc{Full+SP+RT+TA}& 66.3\scriptsize{$\pm0.68$} & 59.1\scriptsize{$\pm0.86$} & 56.1\scriptsize{$\pm0.78$} & 91.9\scriptsize{$\pm3.00$} 
%& 62.4\scriptsize{$\pm0.81$} & 53.4\scriptsize{$\pm4.15$} &27.4\scriptsize{$\pm1.10$} &38.5\scriptsize{$\pm1.28$}\\
\hline
% \hline
% \underline{\textbf{ Pretrain+Multi-task}} & & &&& & & &\\
% \textsc{BiLSTM}& 67.9\scriptsize{$\pm0.50$} & 
% \textbf{61.8}\scriptsize{$\pm0.74$} & \textbf{58.6}\scriptsize{$\pm0.70$} &  \textbf{71.9}\scriptsize{$\pm0.29$}
% & 62.9\scriptsize{$\pm0.24$} &59.2\scriptsize{$\pm5.24$} & 29.8\scriptsize{$\pm1.83$} &  \textbf{39.6}\scriptsize{$\pm0.42$}\\
% \hline
\end{tabular}
}
\end{center}
% \vskip -1em
\caption{\label{tab:main} Main comparison results displayed with average scores (in \%) and their standard deviations over the
results with 5 sets of random initialization seeds. The best results in each column are in \textbf{bold}. Our model yields better scores than all comparisons for all metrics.
}
% \vskip -0.5em
\end{table*}
Table \ref{tab:main} reports the main results on the two datasets. Several interesting observations can be drawn:

$\bullet$~\textit{History and attention mechanism are useful.}
Compared to \textsc{BiGRU}, both \textsc{BiGRU+History} and \textsc{BiGRU+Att} achieve better performance.
The integration of them, i.e., \textsc{Full Main}, brings greater improvement, which means both user's past behaviors and the key turns of current context are important to signal the user's re-entry behavior.

$\bullet$~\textit{Self-supervised tasks are all beneficial.} 
The main model trained with any one of the three self-supervised tasks outperforms the main model itself.
% The three self-supervised tasks all yield better performance than main model without any auxiliary tasks.
Specifically, TA task achieves the best performance on AUC and F1 on both datasets. 
%This demonstrates the effectiveness of both spread pattern and repeated pattern in re-entry prediction.

$\bullet$~\textit{Self-supervised methods perform better than BERT-based model.} 
% Although BERT shows the compelling performance in many tasks that are in formal language, it's not good at re-entry prediction that based on informal language in social media. 
Compared to \textsc{BERT+BiLSTM}, \textsc{Full Main} trained with SP or RT task achieves comparable performance on Twitter and better performance on Reddit.
Besides, \textsc{Full Main} trained with TA task consistently outperforms \textsc{BERT+BiLSTM} on both datasets. 
%These results further validate the effectiveness of our devised self-supervised tasks, especially TA task.
% show introducing our devised self-supervised tasks in the training can significantly improve the performance of re-entry prediction model. 
% , which further validate that .
%Although BERT can learn meaningful sentence representations, it is not very suitable for our task which is based on social media and needs understanding of personality. 
% Our task-oriented tasks, instead, are concerned on users behaviours and show better performance.
The reason might be that the TA task can better capture the user's re-entry behaviors and thus leads to better performance of the main model.

$\bullet$~\textit{Self-supervised learning can make the performance more stable.} We can see that all models with auxiliary self-supervised tasks have a smaller standard deviation, which means self-supervised learning can reduce the impact of the model's parameter initialization and make the performance more stable.

%$\bullet$~\textit{Re-entry Prediction for Reddit is more challenging.} All models perform no more than $40$ F1-scores in Reddit dataset. This is because Reddit dataset is rather imbalanced where much more users do not come back later (see the last row of Table~\ref{statistics-table1}). It is therefore more difficult for models to learn user's re-entry behavior in it.

% Since large proportion of users in Reddit will not return to previous conversations (see the last row of Table~\ref{statistics-table1}), this shows the difficulty of understanding why users abandon previous discussions with only context information.

\subsection{Effects of Our Self-Supervised Tasks} \label{ssec:comparison_in_detail}
% We have shown our effectiveness in main results in Section \ref{ssec:exp:main_results}. 
To further validate the effects of our self-supervised tasks, we compare them with three generic self-supervised tasks. 
% We first compare them with other generic self-supervised methods. 
Also, we investigate the training efficiency of our main model. 
%with self-supervised learning, main model without self-supervised learning and baselines.

\paragraph{Compare with other self-supervised tasks.}
\begin{table}[t]\setlength{\tabcolsep}{1.2mm}\small
\newcommand{\tabincell}[2]{\begin{tabular}{@{}#1@{}}#2\end{tabular}}
\begin{center}
\scalebox{0.9}{
\begin{tabular}{|l|rrrr|rrrr|}
\hline 
\multirow{2}{*}{Tasks} 
&\multicolumn{4}{c|}{ \tabincell{c}{\textbf{Twitter}} }
&\multicolumn{4}{|c|}{ \tabincell{c}{\textbf{Reddit}} }

\\
\cline{2-9}
& AUC  & Acc & Pre & F1 & AUC &  Acc & Pre & F1 \\
\hline
\hline
\textsc{Replace} & 65.6 & 58.2 & 55.5  &69.3
& 62.3 & 54.9 & 28.8 &39.2\\
\textsc{Switch} & 65.8& 58.0 & 56.8  &69.1
& 61.1 & 52.8 & 27.2 &38.9\\
\textsc{Mask} & 64.3 & 57.5 & 55.3  &68.5
& 60.7 & 53.0 & 27.6 &38.7\\
\textsc{TA (Our)} & \textbf{68.6} & \textbf{61.0} & \textbf{58.4}  &\textbf{70.5}
&\textbf{64.6} & \textbf{57.7} & \textbf{29.1} & \textbf{40.6}\\
\hline
\end{tabular}
}
\end{center}
% \vskip -1em
\caption{\label{tab:difftask} Comparison results (in \%) of different self-supervised tasks. TA task yields better performance than Replace, Switch and Mask on all metrics.
}

\end{table}
We compare our best task TA with three popular self-supervised tasks: \textbf{Replace}, \textbf{Switch} and \textbf{Mask}. We follow the turn-level setting in \citet{wang2019self} and implement them as follows. 
\textbf{Replace:} randomly replaces some turns in a conversation with random turns from other conversations, then predict which turns are replaced (each turn has one label, while $1$ means replaced, $0$ otherwise). 
\textbf{Switch:} randomly switches some turns of the conversation, then predict which turns are not in the original positions ($1$ means not in the original position, otherwise $0$). 
\textbf{Mask:} randomly masks the representations of some turns, then predicts them from a candidate list. 
%\paragraph{Comparisons of performance.}
Refer to Table \ref{tab:difftask}, our self-supervised task outperforms other generic tasks on the both datasets. This is probably because our tasks can capture more useful information (e.g., thread pattern and user trajectory) which are vital to re-entry prediction. 
%A more detailed analysis is in Section \ref{}.

\paragraph{Compare with baselines.}
\begin{figure}[t]
\centering
\subfigure[Convergent Speed]{\label{sfig:exp:convergence}
\includegraphics[width=3.8cm]{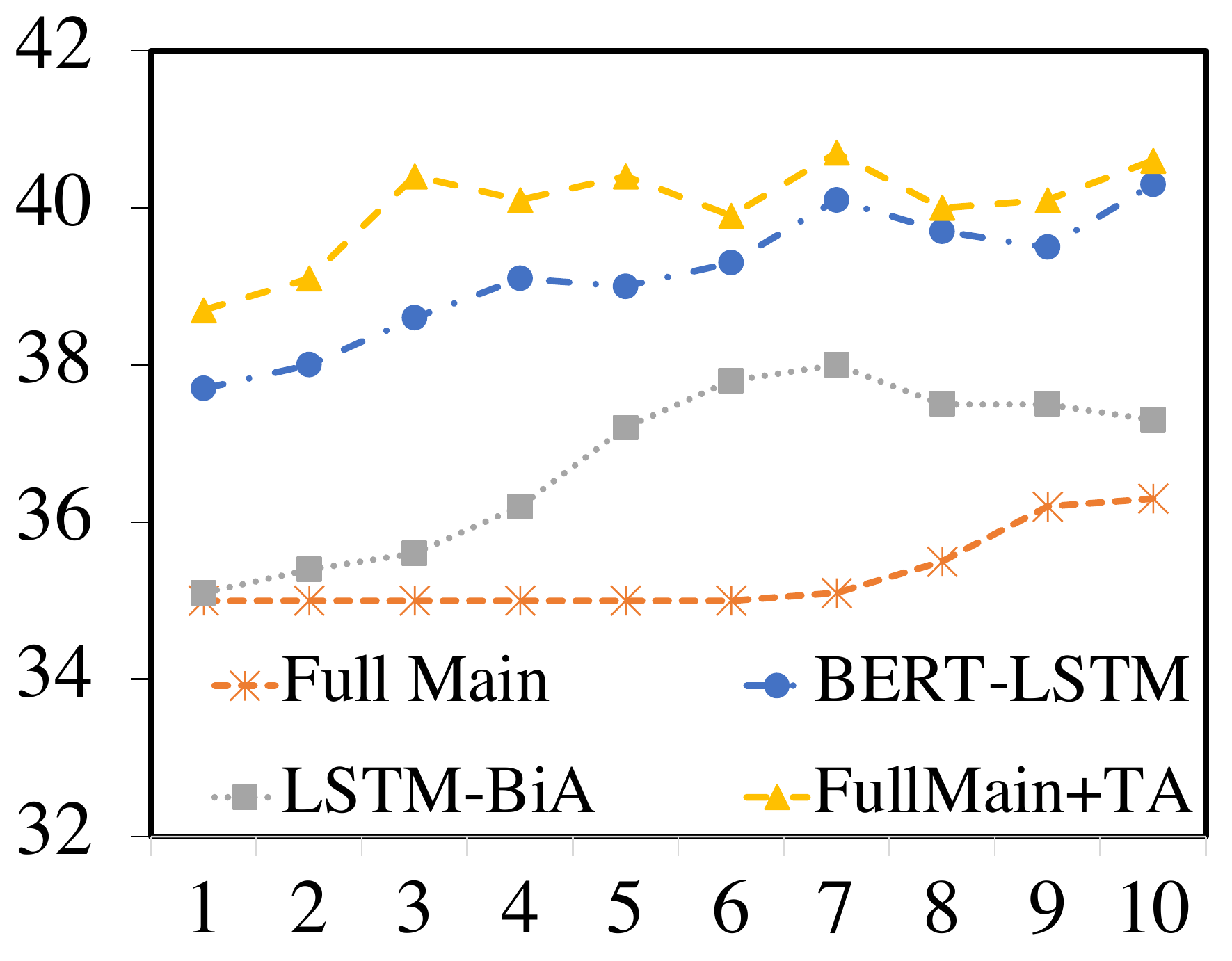}
}
\subfigure[Para Size \& Train Time] {\label{sfig:exp:size_time}
\includegraphics[width=3.4cm]{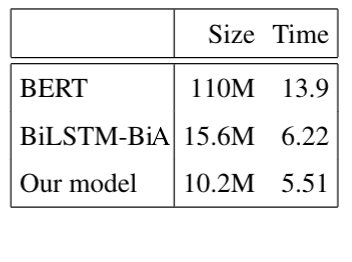}
}
\vskip -0.5em
\label{fig:none-self-supervised}
\caption{
For Fig. \ref{sfig:exp:convergence}, X-axis: epoch index, Y-axis: F1 score. For Tab. \ref{sfig:exp:size_time}, ``Size" means the parameter size, ``Time" refers to the time that one epoch needed. 
}
\vskip -1em
\end{figure}
As discussed in Section \ref{ssec:exp:main_results}, models with self-supervised learning show more stable performance. 
We further explore their differences with respect to convergence speed, parameter size, and training time. 
We present F1 scores (in validation set) of the first 10 epochs for the four models, \textsc{BERT+BiLSTM}, \textsc{BiLSTM-BiA} \cite{zeng-etal-2019-joint}, our main model without self-supervised learning (\textsc{Full Main}) and our main model with self-supervised learning (\textsc{FullMain+TA}) in Figure \ref{sfig:exp:convergence}. 
As we can see, \textsc{FullMain+TA} achieves the highest F1 scores from the first epoch.
% , converges much faster than other models. 
This is due to the benefits of our efficient pattern-guided self-supervised learning. 
On the other hand, \textsc{FullMain+TA} converges early at around the third epoch while the other models are trained slowly and converge later.
% at around the ninth epoch. 
We also present the parameter size and train time of one epoch for \textsc{BERT-BiLSTM} (BERT), \textsc{BiLSTM-BiA} and \textsc{FullMain+TA} (Our model) in Table \ref{sfig:exp:size_time}. It can be seen that \textsc{Full Main+TA} has fewer parameters and faster training speed. 
% We also compare our model with other models without self-supervised learning, such as BiLSTM+BiA \cite{zeng-etal-2019-joint} and our main model together with our main model with self-supervised tasks RA. BERT \cite{devlin2018bert} We first compare convergent speed of them. 
% As shown in Figure \ref{sfig:exp:convergence}, we present the F1 scores (in validation set) of the first 10 epochs of each model in Figure \ref{sfig:exp:convergence}. As can be seen, Our model convergent at around the third epoch while the two model without pretraining, LSTM-BiA and BiLSTM, are convergent slowly and do not convergent in the range of 10 epochs. BERT can be regarded as a model that has pretrained in large-scale datasets. BERT can convergent at the seventh epoch. We can also find that models (i.e., AR and BERT) with pretraining have a better start then those without pretraining. Because 
%To better illustrate the benefits of self-supervised tasks, we discuss two aspect of advantages, convergent speed and .

\subsection{How Do Our Methods Work?}\label{ssec:why}
We also explore the inherent properties of our methods and show how they work. In this way, we would like to point out some key ideas in designing task-oriented self-supervised tasks.
% To figure out which kinds of instances are benefited by our self-supervised tasks and how they improve them, we first show the performance improvement of different user patterns after applying our self-supervised tasks. Then we discuss the results when the labeling for self-supervised tasks is opposite. 

\paragraph{What types of conversations are benefited?}
\begin{figure}[t]
\centering
\includegraphics[width=0.45\textwidth]{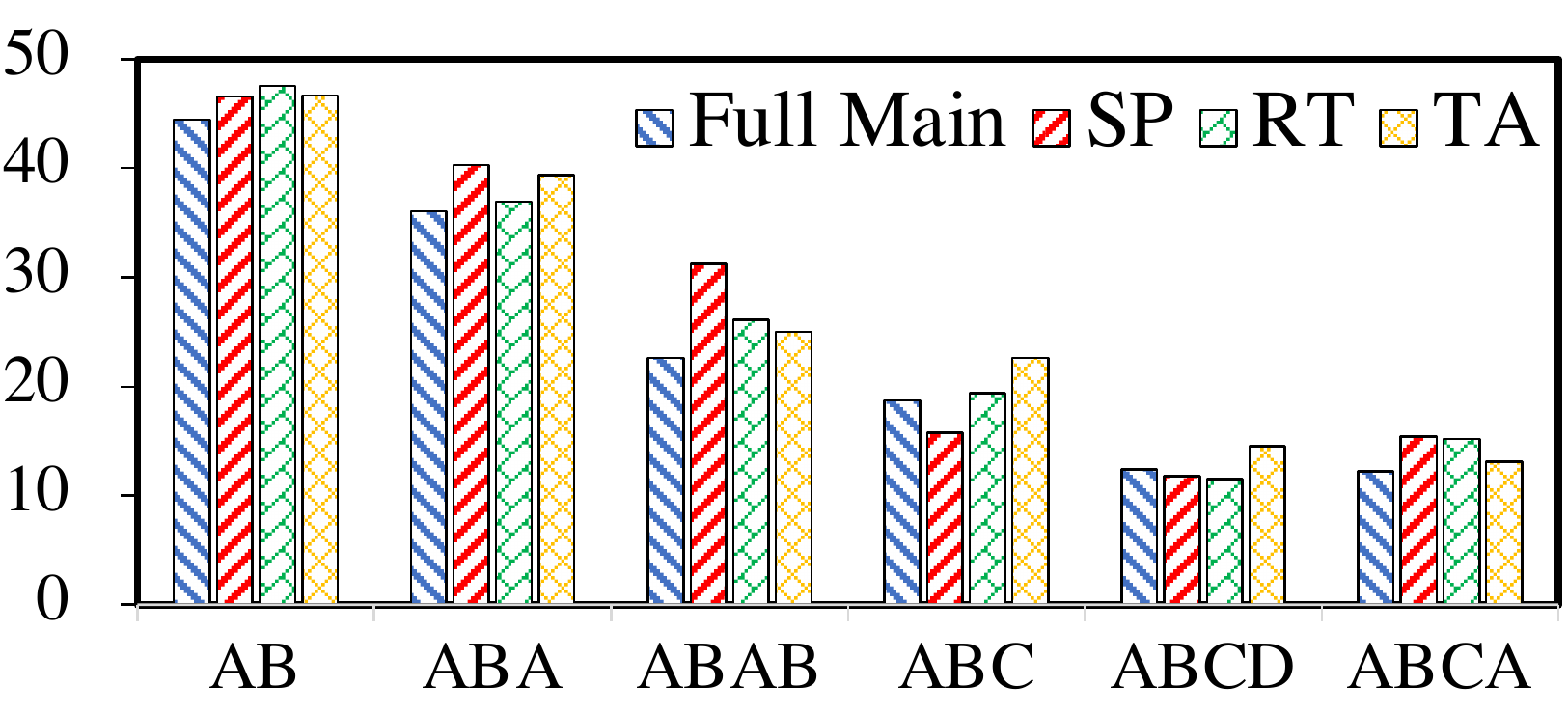}
\caption{\label{fig:exp:pattern_performance}  X-axis: thread pattern (the same meaning as in Fig. \ref{fig:statistis:pattern_rate}). Y-axis: F1 score (in \%). 
%``Tn'' in X-axis means the $n^{th}$ turn of the conversation.
} 
% \vskip -1.0em
\end{figure}

To understand how our self-supervised tasks work, we explore the performance of six different kinds of conversations categorized by their thread patterns. Three of them (``AB'', ``ABA'' and ``ABAB'') are focused conversations, the others (``ABC'', ``ABCD'' and ``ABCA'') are expansionary conversations. 
The results together with our main model without self-supervised (\textsc{Full Main}) are displayed in Figure \ref{fig:exp:pattern_performance}. It can be seen that SP brings larger gains for focused conversations than expansionary ones; and RT improves the cases with repeated target users (``ABA'', ``ABAB'' and ``ABCA'') most. Such results show that SP and RT tasks benefit the main task by improving performance on their positive instances. 
%, which are just what they are good at. 
This raises a suggestion on designing task-oriented self-supervised tasks, i.e., \textit{choosing tasks related to the instances that your current model is not good at}.
Also, different self-supervised tasks can be proposed for different purposes in a real system. 
On the other hand, TA performs consistently better in all six cases, because the turn-level labeling emphasizes the model capability of tackling all kinds of conversations.
This raises another suggestion, i.e., \textit{designing tasks that can reflect model's ability in different dimensions}.

% the gained benefits for different data type, we explore self-supervised tasks performance on six conversation user patterns, three of them ("AB", "ABA" and "ABAB") are focused conversations, the others ("ABC", "ABCD" and "ABCA") are expansionary conversations. The Results is in Figure \ref{fig:statistis:pattern_performance}. It can be seen that focused conversations has better F1 scores than expansionary conversations and our models with all kinds of  self-supervised learning outperforms than Bi-LSTM. In expansionary conversations prediction, TA tasks outperforms Bi-LSTM.

\paragraph{Will our methods still work if the labels are inverted?} In general, when we evaluate the performance of a task, positive instances count more than negative instances, since we care more about true positives in calculating precision and recall. Therefore, we wonder whether the labeling strategies will affect the results. To this end, we invert the labels of our self-supervised tasks by changing the label 1 to 0 and label 0 to 1. For example, we used to label focused conversations as 1 and expansionary conversations as 0 in SP task. Now we label them with opposite labels to explore how the methods work. From the results shown in Figure \ref{sfig:exp:inverse}, the performance of inverted SP and inverted RT is even poorer than the model without self-supervised tasks (W/O). Inverted TA shows better performance than W/O, but the F1 score is still lower than the original labeled TA. 
We attribute such performance drop to the inconsistent labeling between auxiliary task and main task. This means that the positive label in the auxiliary task should be related to that in the main task so as to enhance learning.
Therefore, we turn to the final finding in our experiments, i.e., \textit{labeling strategies make a difference for the designed self-supervised tasks}.

\subsection{Further Discussion} \label{ssec:more_discuss}
%\paragraph{Effects of repeated engagement}
\paragraph{Effects of user history.}
\begin{figure}[t]
\centering
\subfigure[Effects of Labeling  ]{\label{sfig:exp:inverse}
\includegraphics[width=3.6cm]{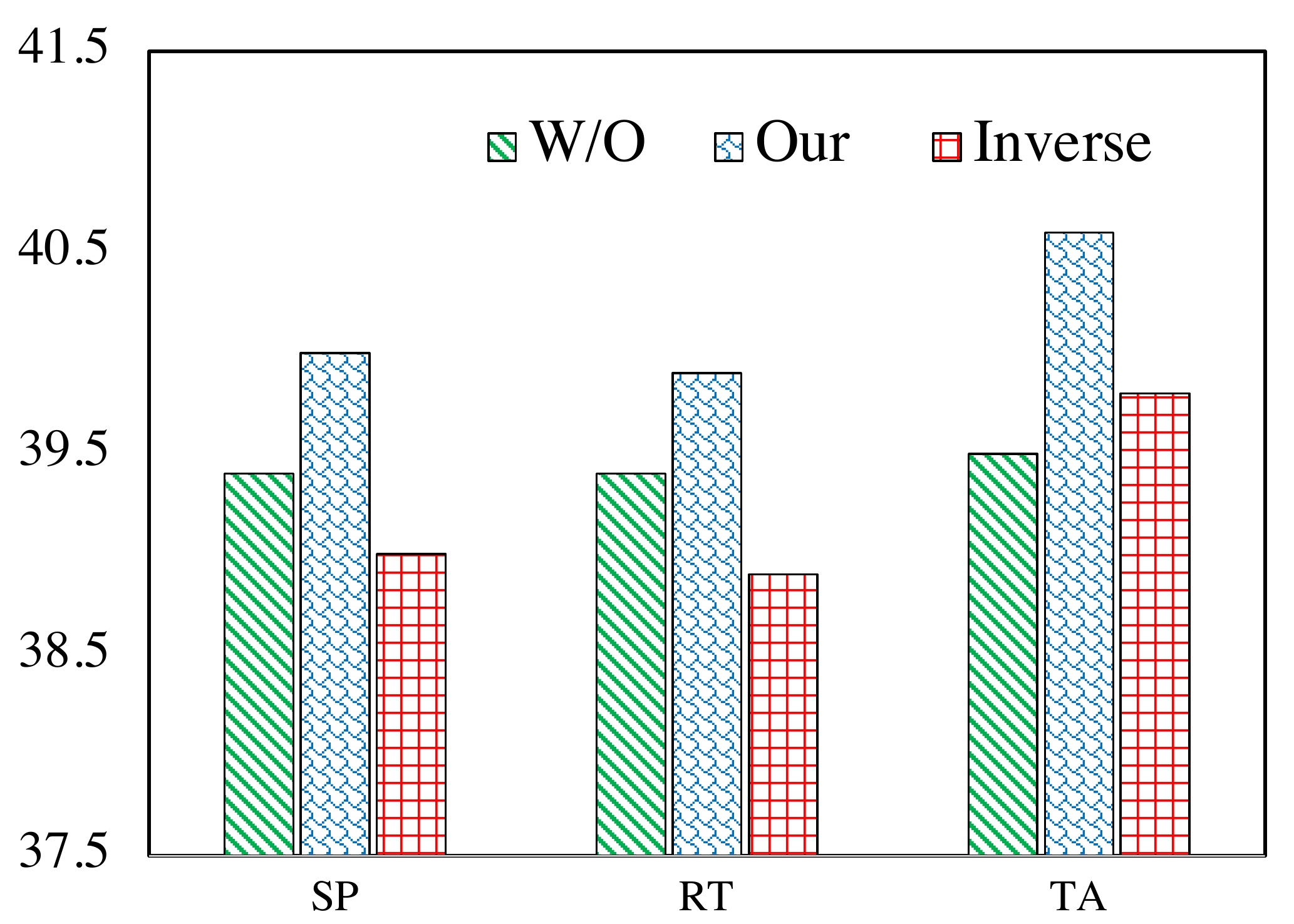}
}
\subfigure[Effects of History Num]{\label{sfig:exp:history}
\includegraphics[width=3.6cm]{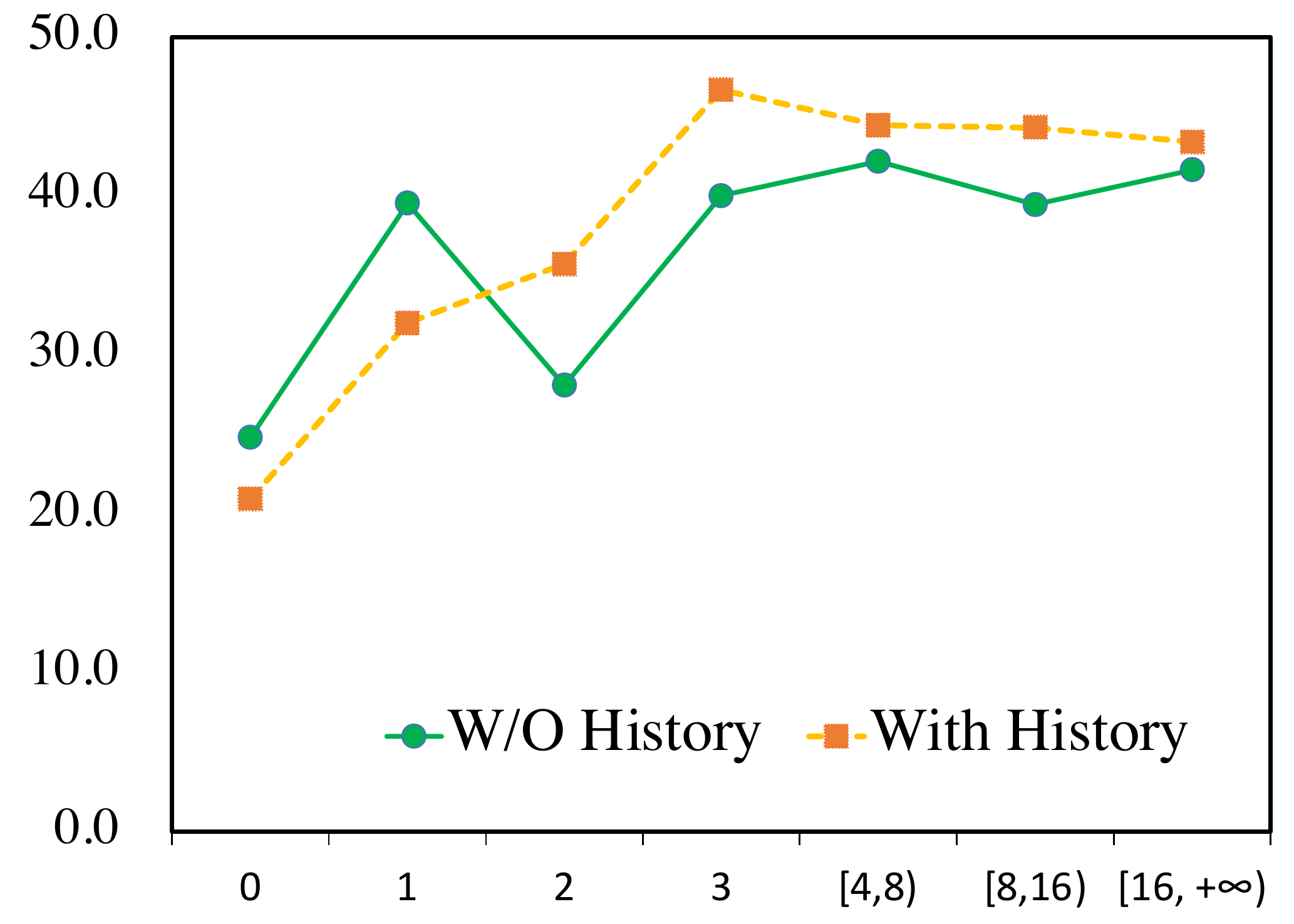}
}
% \vskip -0.5em
\label{fig:inverse-hist}
\caption{ Fig.\ref{sfig:exp:inverse} displays the F1 scores (in \%, Y-axis) for SP, RT and TA in three different scenarios: without these tasks (W/O), the labels for these tasks are the same as main results (Our) and the labels are inverted (Inverse). For Fig.\ref{sfig:exp:history}, X-axis: the number of history turns that target user has, Y-axis: F1 score (in \%). 
}
% \vskip -0.5em
\end{figure}
To understand how user history affects prediction, we present F1 scores for the model with history and without history in Figure \ref{sfig:exp:history}. Our model with history performs better for users having more than 1 history conversation and perform worse in the cases of only 0 and 1. 
This is because our model needs sufficient information to capture personalized features.
% The reason can be model cannot learn well with few history.

\paragraph{Error analysis.}
We have tried the joint training of all three auxiliary tasks and find that performance is similar to training only with TA task. This might be attributed to the difficulty of balancing among so many tasks during joint training. Another reason is that TA task has already covered the information in SP and RT task as its idea comes from the combination of the previous two tasks.
On the other hand, our model performs worse in predicting the expansionary conversations (Figure~\ref{fig:exp:pattern_performance}), since most users in such conversations tend to not return, and the reasons for that might be diverse, e.g., too busy to reply. We leave how to enhance the performance in such cases as our future work.

% \paragraph{Multi-task learning v.s. Pre-training.}
% In our main comparison results (Section \ref{ssec:exp:main_results}), we train the self-supervised tasks in the manner of pretraining plus multi-task learning. Here we explore their effects separately, i.e., we train the self-supervised tasks in pretraining or multi-task manner. Table \ref{tab:multitask} shows the results. As can be seen, our model performs worse with only one kind of manner. And it’s hard to say which training method is better, although pretraining is more stable.
% \input{tables/multi_task}

% \paragraph{Joint effects of self-supervised tasks.}
% \input{tables/joint_self-supervised}
% We have shown the results of three self-supervised tasks separately in Table \ref{tab:main}, here we further exploit their joint effects by training them together and results are in Table \ref{tab:joint-self-supervised}. Surprisingly, the joint training of them does not achieve better results. The reason can be two,

\section{Conclusion}\label{sec:conclusion}
We present a basic model with three novel self-supervised tasks for re-entry prediction. Experiments on two newly constructed conversation datasets, Twitter and Reddit, show that our model outperforms the previous models with fewer parameters and faster convergence. Further discussions provide more insights on how our model works and how to design task-oriented self-supervised tasks.

\section*{Acknowledgements}
The research described in this paper is partially supported by HK GRF \#14204118 and HK RSFS \#3133237. We thank the three anonymous reviewers for the insightful suggestions on various aspects of this work.
\bibliography{anthology,custom}
\bibliographystyle{acl_natbib}

% \appendix

% \section{Example Appendix}
% \label{sec:appendix}

% This is an appendix.

\end{document}